\documentclass{article}
\usepackage{arxiv}

\usepackage[utf8]{inputenc} 
\usepackage[T1]{fontenc}    
\usepackage{hyperref}       
\usepackage{url}            
\usepackage{booktabs}       
\usepackage{amsfonts}       
\usepackage{amsmath}
\usepackage{nicefrac}       
\usepackage{microtype}      
\usepackage{lipsum}
\usepackage{graphicx}
\usepackage{float}
\usepackage{tabularx}
\usepackage{hhline}
\title{Towards Unsupervised Cancer Subtyping: Predicting Prognosis Using A Histologic Visual Dictionary}

\usepackage{authblk}

\author[1,2\thanks{\texttt{muhammah@mskcc.org, http://thomasfuchslab.org/hm}}]{Hassan Muhammad}
\author[1]{Carlie S. Sigel}
\author[1,2]{Gabriele Campanella}
\author[3]{Thomas Boerner}
\author[3]{Linda M. Pak}
\author[4]{Stefan Büttner}
\author[4]{Jan N.M. IJzermans}
\author[4]{Bas Groot Koerkamp}
\author[5]{Michael Doukas}
\author[3]{William R. Jarnagin}
\author[3]{Amber Simpson}
\author[1,2]{Thomas J. Fuchs}
\affil[1]{Department of Pathology | Memorial Sloan-Kettering Cancer Center}
\affil[2]{Department of Physiology, Biophysics, and Systems Biology | Weill Cornell Medicine}
\affil[3]{Department of Surgery | Memorial Sloan-Kettering Cancer Center}
\affil[4]{Department of Surgery | Erasmus Medical Center-Rotterdam}
\affil[5]{Department of Pathology | Erasmus Medical Center-Rotterdam}

\begin{document}
\maketitle

\begin{abstract}
Unlike common cancers, such as those of the prostate and breast, tumor grading in rare cancers is difficult and largely undefined because of small sample sizes, the sheer volume of time needed to undertake on such a task, and the inherent difficulty of extracting human-observed patterns. One of the most challenging examples is intrahepatic cholangiocarcinoma (ICC), a primary liver cancer arising from the biliary system, for which there is well-recognized tumor heterogeneity and no grading paradigm or prognostic biomarkers. In this paper, we propose a new unsupervised deep convolutional autoencoder-based clustering model that groups together cellular and structural morphologies of tumor in 246 ICC digitized whole slides, based on visual similarity. From this visual dictionary of histologic patterns, we use the clusters as covariates to train Cox-proportional hazard survival models. In univariate analysis, three clusters were significantly associated with recurrence-free survival. Combinations of these clusters were significant in multivariate analysis. In a multivariate analysis of all clusters, five showed significance to recurrence-free survival, however the overall model was not measured to be significant. Finally, a pathologist assigned clinical terminology to the significant clusters in the visual dictionary and found evidence supporting the hypothesis that collagen-enriched fibrosis plays a role in disease severity. These results offer insight into the future of cancer subtyping and show that computational pathology can contribute to disease prognostication, especially in rare cancers.
\end{abstract}


\section{Introduction}
Cancer grading is an important tool used to predict disease prognosis and direct therapy. Commonly occurring cancers, such as those of breast and prostate, have well established grading schemes validated on large sample sizes \cite{bloom1957histological}. The grading system for prostate cancer is the foremost internationally accepted grading scheme in cancer medicine. This system, known as the Gleason Grading System (GGS), developed by Donald Gleason, related observed histologic patterns in pathology to outcome data using thousands of cases. After nearly twenty years of repeatedly designing and validating his prognostic classifications, he published a final review of grading prostate cancer in 1992. Since then, it has been a clinical standard \cite{delahunt2012gleason}, barring minor revisions in 2005, and again in 2011, by an international society of experts. Although the gold standard in prostate cancer stratification, GGS is subject to ongoing critical assessment. In 2016, McKenney et al. created a new dictionary of histologic patterns, independent from those which constituted the GGS, from a cohort of 1275 cases followed over five years. Using these specific patterns, additional risk groups were determined within GGS grades, supporting the notion that GGS can be further optimized \cite{mckenney2016histologic}. The manual labor required to identify different histologic patterns and use them to stratify patients into different risk groups is an extremely complex task requiring years of effort and repeat reviews of large amounts of visual data, often by one pathologist. The story of prostate carcinoma grading indicates different observers may identify different or incomplete sets of patterns. 

Developing a grading system for a rare cancer poses a unique set of challenges. Intrahepatic cholangiocarcinoma (ICC), a cancer of the bile duct, has an incidence of approximately 1 in 100,000 in the United States, and rising \cite{saha2016forty}. Currently, there exists no universally accepted histopathology-based subtyping or grading system for ICC and studies classifying ICC into different risk groups have been inconsistent \cite{nakajima1988histopathologic, aishima2007proposal, sempoux2011intrahepatic}.  A major limiting factor to subtyping ICC is that only small cohorts are available to each research institution. Our institution recently published a study  \cite{sigel2018intrahepatic} using one of the world's largest cohorts of ICC (n = 184), expanding one proposed histology-based binary subtyping \cite{hayashi2016distinct} into four risk groups but still found no significant association with survival. 

There is an urgent need for efficient identification of prognostically relevant cellular and structural morphologies from limited histology datasets of rare cancers to build risk stratification systems which are currently lacking across many cancer types. Ideally, these systems should utilize a visual dictionary of histologic patterns that is comprehensive and reproducible. Once generated, such a visual dictionary must be translatable to histopathological terms universally understood by pathologists. Computational pathology \cite{fuchs2011computational} offers a new set of tools, and more importantly, a new way of approaching the historical challenges of subtyping cancers using computer vision-based deep learning, leveraging the digitization of pathology slides, and taking advantage of the latest advances in computational processing power. In this paper, we offer a new deep learning-based model which can create such a visual dictionary and show utility by stratifying ICC, based on morphology at the cellular level for the first time.

\section{Materials and Methods}
\label{sec:headings}

Cancer histopathology images exhibit high intra- and inter-heterogeneity because of their size (as large as tens of billions of pixels). Different spatial or temporal sampling of a tumor can have sub-populations of cells with unique genomes, theoretically resulting in visually different patterns of histology \cite{bedard2013tumour}. In order to effectively cluster this extremely large amount of high intra-variance data into subsets which are based on similar morphologies, we propose combining a neural network-based clustering cost-function, previously shown to outperform traditional clustering techniques on images of hand-written digits \cite{song2013auto}, with a novel deep convolutional architecture. Finally, we assess the power and usefulness of this clustering model by conducting survival analysis, using both Cox-proportional hazard modeling and Kaplan-Meier survival estimation, to measure if each cluster of histomorphologies has significant correlation to recurrence of cancer after resection.

\subsection{Deep Clustering Convolutional Auto Encoder}
A convolutional auto-encoder is made of two parts, an encoder and decoder. The encoder layers project an image into a lower dimensional representation, an embedding, through a series of convolution, pooling, and activation functions. This is described in equation \ref{eq:embedding}a, where $x_i$ is an input image or input batch of images transformed by $f_\theta()$, and $z_i$ is the resulting representation embedding. The decoder layers try to reconstruct the original input image from its embedding using similar functions. Mean-squared-error loss (MSE) is commonly used to optimize such a model, updating model weights ($\theta$) relative to the error between the original (input, $x_i$) image and the reconstruction (output, $x_i^{'}$) image in a set of $N$ images. This is shown in equation \ref{eq:embedding}b.

\begin{equation}\label{eq:embedding}
    (a)\ z_i = f_{\theta}(x_i) \quad \quad \quad (b)\ \epsilon = \min_{\theta}\frac{1}{N}\sum_{i=1}^{N}||x_i - x_i^{'}||^2 
\end{equation}

Although a convolutional auto-encoder can learn effective lower-dimensional representations of a set of images, it does not cluster together samples with similar morphology. To overcome this problem, we amend the traditional MSE-loss function by using the reconstruction-clustering error function, as proposed first by Song et al. \cite{song2013auto}:

\begin{equation}\label{eq:cost}
\epsilon = \min_{\theta}\frac{1}{N}\sum_{i=1}^{N}||x_i - x_i^{'}||^2 + \lambda\sum_{i=1}^{N}||z_i-c_i^{*}||^2,
\end{equation}

where $z_i$ is the embedding as defined in equation \ref{eq:embedding}a, $c^*_i$ is the centroid assigned to sample $x_i$ from the previous training epoch, and $\lambda$ is a weighting parameter. Cluster assignment is determined by finding the shortest Euclidean distance between a sample embedding from epoch $t$ and a centroid, across $j$ centroids from epoch $t-1$: 

\begin{equation}\label{eq:nearest_center}
c_i^{*} = \operatorname*{arg\,min}_{j} || z_i^t - c_j^{t-1}||^2
\end{equation}

The algorithm is initialized by assigning a random cluster to each sample. Centroid locations are calculated for each cluster class by equation \ref{eq:calc_centers}. Each mini-batch is forwarded through the model and network weights are respectively updated. At the end of an epoch, defined by the forward-passing of all mini-batches, cluster assignments are updated by equation \ref{eq:nearest_center}, given the new embedding space. Finally, the centroid locations are updated from the new cluster assignments. This process is repeated until convergence. Figure \ref{fig:model} shows a visualization of this training procedure.

\begin{equation}\label{eq:calc_centers}
c_j^{t} = \frac{\sum_{t=1}^{N}z_i}{|C_j^{t-1}|}
\end{equation}

\subsection{Dataset}
Whole slide images were obtained from Memorial Sloan Kettering Cancer Center (MSKCC) and Erasmus Medical Center with approval from each respective Institutional Review Boards. In total, 246 patients with resected ICC without neoadjuvant chemotherapy were included in the analysis. All slides were digitized at MSKCC using Aperio AT2 scanners (Leica Biosystems; Wetzlar Germany). Up-to-date retrospective data for recurrence free survival after resection was also obtained. Though currently a small sample size, this collection is the largest known retrospective ICC dataset in the world.

\begin{figure}[H]
  \centering
  \includegraphics[width=\linewidth]{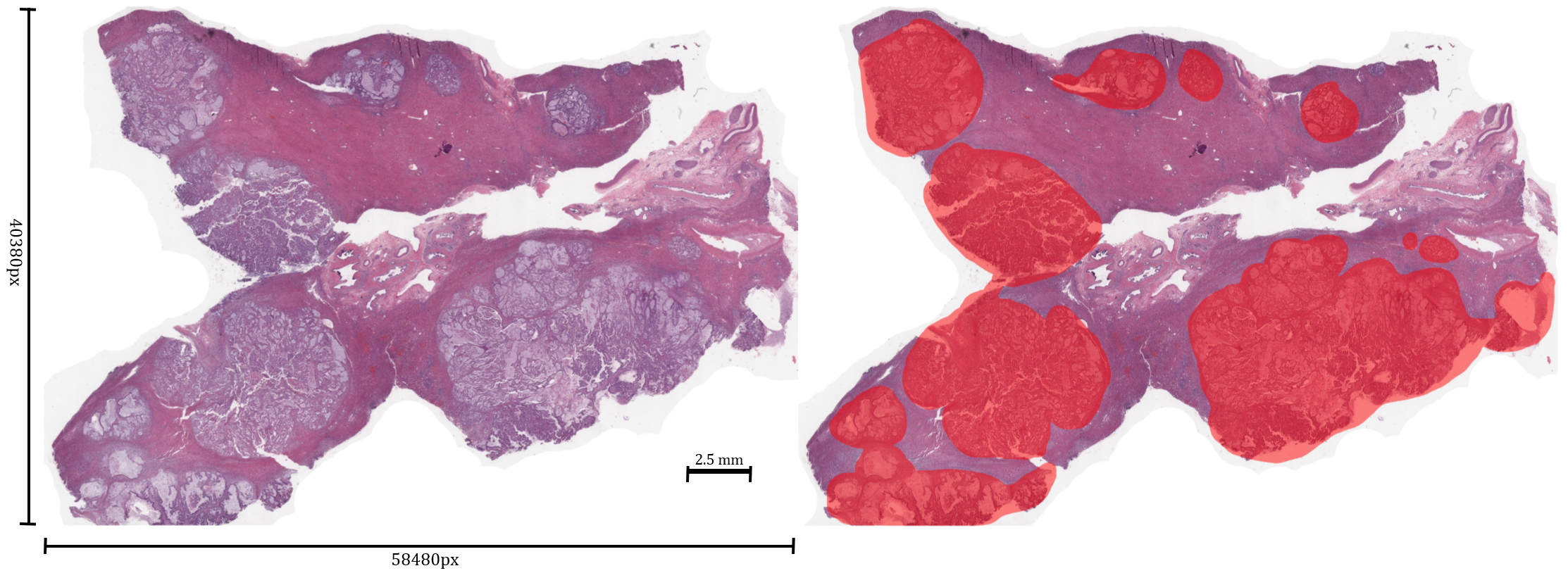}
  \caption{(Left) An example of a digitized slide at low magnification. These slides are quite large, as this average example comprises at least one billion pixels of tissue. (Right) The same slide with regions of tumor annotated in red. Note that annotation may cover areas of background. Our novel tiling protocol with quality control ensures that tiles contain high resolution and clear images of tissue.}
  \label{fig:labeling}
\end{figure}

A library of extracted image tiles was generated from all digitized slides. First, each slide was reduced to a thumbnail, where one pixel in the thumbnail represented a 224x224px tile in the slide at 20x magnification. Next, using Otsu thresholding on the thumbnail, a binary mask of tissue (positive) vs. background (negative) was generated \cite{campanella2018terabyte}. Connected components below 10 thumbnail pixels in tissue were considered background to exclude dirt or other insignificant masses in the digitized slide. Finally, mathematical morphology was used to erode the tissue mask by one thumbnail pixel to minimize tiles with partial background. To separate the problem of cancer subtyping, as discussed in this paper, from the problem of tumor segmentation, the areas of tumor were manually annotated using a web-based whole slide viewer. Using a touchscreen (Surface Pro 3, Surface Studio; Microsoft Inc., Redmond, WA, USA), a pathologist painted over regions of tumor to identify where tiles should be extracted. Figure \ref{fig:labeling} illustrates an example of this annotation. Tiles were added to the training set if they lay completely within these regions of identified tumor.

\subsubsection{Quality Control}
\begin{figure}[H]
  \centering
  \includegraphics[width=\linewidth]{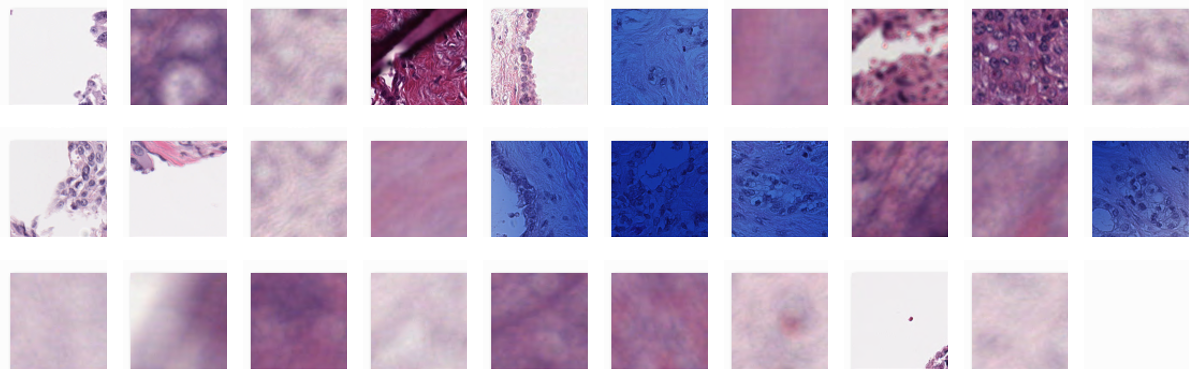}
  \caption{Randomly selected examples of tiles excluded by the quality control algorithm. Interestingly, this method also helped identify tiles with ink stains, folds, and those which include partial background, tiles which the first stage of tile generation was designed to exclude as much as possible.}
  \label{fig:excluded}
\end{figure}

Scanning artifacts such as out-of-focus areas of an image can impact model performance on smaller datasets. A deep convolutional neural network was trained to detect blurred tiles to further reduce noise in the dataset. Training a detector on real blur data was beyond the scope of this study because obtaining annotations for blurred regions in the slide is unfeasible and would also create a strong class imbalance between blurred and sharp tiles. To prepare data for training a blur detector, we used an approach similar to a method described in \cite{campanella2018towards}. To start, half of the tiles were artificially blurred by applying a Gaussian-blur filter with a random filter radius ranging from 1 to 10. The other half were labeled "sharp" and no change was made to them. A ResNet18 was trained to output an image quality score by regressing over the values of the applied filter radius using MSE. A value of 0 was used for images in the sharp class. Finally, a threshold was manually selected to exclude blurred images based on the output value from the detector. Figure \ref{fig:excluded} shows randomly selected examples of tiles excluded based on the blur detector. 

\subsection{Architecture and Training}
We propose a novel convolutional autoencoder architecture to optimize performance in image reconstruction. The encoder is a ResNet18 \cite{he2016deep} which was pretrained on ImageNet \cite{ILSVRC15}. The parameters of all layers of the encoder updated when training the full model on pathology data. The decoder is comprised of five convolutional layers, each with a padding and stride of 1, for keeping the tensor size constant with each convolution operation. Upsampling is used before each convolution step to increase the size of the feature map. Empirically, batch normalization layers did not improve reconstruction performance and thus, were excluded.

\begin{figure}[H]
  \centering
  \includegraphics[width=\linewidth]{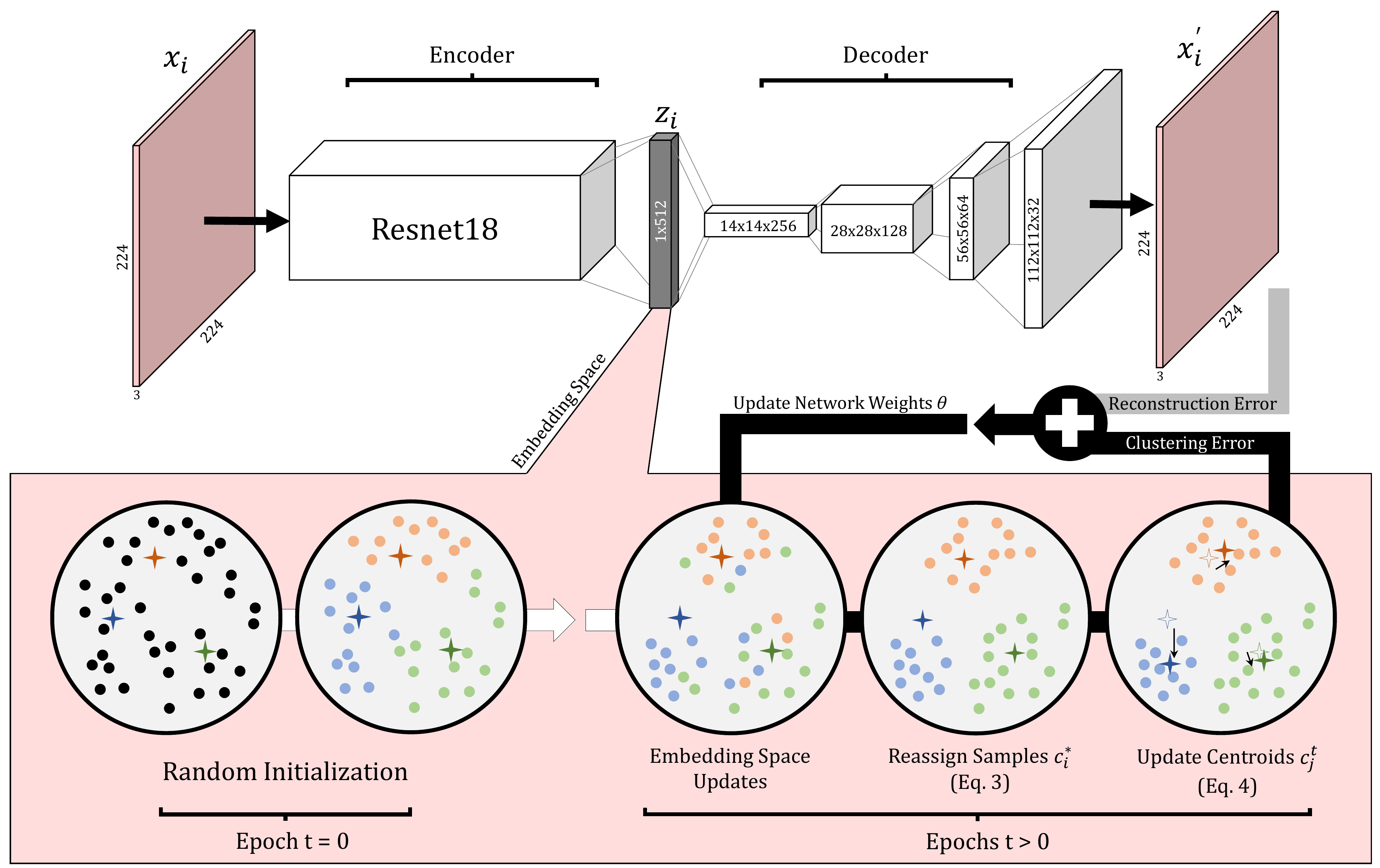}
  \caption{At each iteration, the model is updated in two steps. After a each forward-pass of a minibatch, the network weights are updated. At the end of each epoch, centroid locations are updated by reassigning all samples in the newly updated embedding space to the nearest centroid from the previous epoch, as described in equation \ref{eq:nearest_center}. Finally, each centroid location is recalculated using equation \ref{eq:calc_centers}. All centroids are randomly initialized before training.}
  \label{fig:model}
\end{figure}

Two properties of the model need to be optimized: first, the weights of the network, $\theta$, and then locations of the cluster centers, or centroids, in the embedding space, $C_j$. In order to minimize equation \ref{eq:cost} and update $\theta$, the previous training epoch's set of centroids, $C_j^{t-1}$, is used. In the case of the first training epoch, centroid locations are randomly assigned upon initialization. A training epoch is defined by the forward-passing of all mini-batches once through the network. After $\theta$ have been updated, all samples are reassigned to the nearest centroid using equation \ref{eq:nearest_center}. Finally, all centroid locations are updated using equation \ref{eq:calc_centers} and used in the calculations of the next training epoch. Figure \ref{fig:model} illustrates this process and architecture.

All training was done on DGX-1 compute nodes (NVIDIA Corp., Santa Clara, CA) using PyTorch 0.4 on Linux CentOS 7. The model was trained using Adam optimization for 125 epochs, a learning rate of $1e^{-2}$, and weight decay of $1e^{-4}$. The learning rate was decreased by 0.1 every 50 epochs. The clustering weight, $\lambda$, was set to 0.4. Finally, to save on computation time, 500,000 tiles were randomly sampled from the complete tile library to train each model, resulting in approximately 2000 tiles from each slide on average.

\subsubsection{Survival Analysis}
In order to measure the usefulness and effectiveness of the clustered morphological patterns, we conducted slide-level survival analysis, based on which patterns occurred on a given digital slide to its associated outcome data. Each cluster was considered a binary covariate. If one or more tile(s) from a given cluster existed in a slide, the slide was considered positive for the morphological pattern defined by that cluster. A multivariate Cox regression was used to model the impact of all clusters on recurrence based on this binarized representation of each cluster for each patient:

\begin{equation}
    H(t) = h_{o}e^{b_1x_1 + b_2x_2 + ... + b_jx_j},
\end{equation}

where $H(t)$ is the hazard function dependant on time $t$, $h_o$ is a baseline hazard, and covariates $(x_1, x_2,...x_j)$ have coefficients $(b_1, b_2,...b_j)$. A covariate's hazard ratio is defined by $e^{b_j}$. A hazard ratio greater than one indicates that tiles in that cluster contribute to a worse prognosis. Conversely, a hazard ratio less than one contributes to a good prognostic factor. To measure significance in the survival model, p-values based on the Wald statistic are presented for each covariate.

A univariate Cox Regression was also performed on each cluster. Those measured as significant (p < 0.05) were used to build multivariate Cox regressions for each combination. The results are described in Table \ref{tb:univariate}. Finally, we show Kaplan-Meier curves of the prognositically significant clusters by estimating the survival function $S(t)$:

\begin{equation}
    S(t) = \prod_{t_i < t}\frac{n_i - d_i}{n_n},
\end{equation}

where $d_i$ are the number of  recurrence events at time $t$ and $n_i$ are the number of subjects at risk of death or recurrence prior to time $t$. This binary Kaplan-Meier analysis was done for each cluster, and stratification was measured to be significant using a standard Log-Rank Test.  

\subsubsection{Histologic Interpretation}
Clusters measured to be significantly correlated with survival based on the Cox analysis were assigned clinical descriptors by a pathologist using standard histological terminology as shown in table \ref{tb:elements}. For 20 random tiles of each of those clusters, a semi-quantitative assessment of histological elements comprising each tile was recorded. A major feature  was defined as presence of a histological element in >50 percent of a single tile area in greater than 10 tiles of a cluster. Minor features were defined as histological elements in > 50 percent of the tile area in 8-9 tiles of a cluster. A major tumor histology type was defined for a cluster when >50 percent of the tiles containing any amount of tumor were of the same histologic description.

\begin{table}[H]
\centering
\caption{Standard pathology terms were used to build a description of the most common histologic elements appearing in each cluster.}
\label{tb:elements}
\begin{tabular}{@{}lll@{}}
\toprule
Category                      & Histologic Description                       &                               \\ \midrule \midrule   
Debris                        & Granular fibrinoid material, amorphous       &                               \\
                              & Granular fibrinoid material, ghost cells     &                               \\
                              & Granular fibrinoid material, pyknotic nuclei &                               \\
                              & Nectroic tumor                               &                               \\
                              & Red blood cells                              &                               \\ \midrule
Extracellular Matrix          & Collagen, linear fascicles                   &                               \\
                              & Collagen, wavy fascicles                     &                               \\
                              & Collagen, bundles in cross section           &                               \\
                              & Collagen, amorphous                          &                               \\
                              & Mucin                                        &                               \\ \midrule
Hematolymphoid                & Neutrophils                                  &                               \\
                              & Lymphocytes                                  &                               \\
                              & Histiocytes                                  &                               \\ \midrule
Other Non-Neoplastic Elements & Vessel                                       &                               \\
                              & Nerve                                        &                               \\
                              & Hepatocytes                                  &                               \\
                              & Fibroblasts                                  &                               \\ \midrule
Tumor Histology type          & Tubular                                      & High nuclear:cytolasmic ratio \\
                              &                                              & Low nuclear:cytoplasmic ratio \\ \cmidrule(l){2-3} 
                              & Solid                                        & High nuclear:cytolasmic ratio \\
                              &                                              & Low nuclear:cytoplasmic ratio \\ \cmidrule(l){2-3} 
                              & Too limited to classify                      &                               \\ \bottomrule
\end{tabular}
\end{table}

\subsection{Results}
The multivariate Cox model comprising all clusters showed significance in the hazard ratios of clusters 0, 11, 13, 20, and 23 (p < 0.05). Cluster 20 showed a decrease in prognostic risk and clusters 0, 11, 13, and 23 showed an increase in prognostic risk. However, the overall model was not measured to be significant (Likelihood Ratio Test: p = 0.106, Wald Test: p = 0.096, Log-Rank Test: p = 0.076).
 
Clusters 0, 11, and 13 were measured to be significant (p < 0.05) by the univariate Cox Regression, all with a positive influence in prognosis when compared to samples negative for those clusters. Table \ref{tb:univariate} shows the individual univariate hazard ratios for Clusters 0, 11, 13 and all combinations for a multivariate Cox regression when considering only those significant clusters. For these multivariate models, the Wald Test, Likelihood Ratio Test, and Log-Rank Test, all showed significance of p < 0.05.

\begin{figure}[H]
  \centering
  \includegraphics[width=\linewidth]{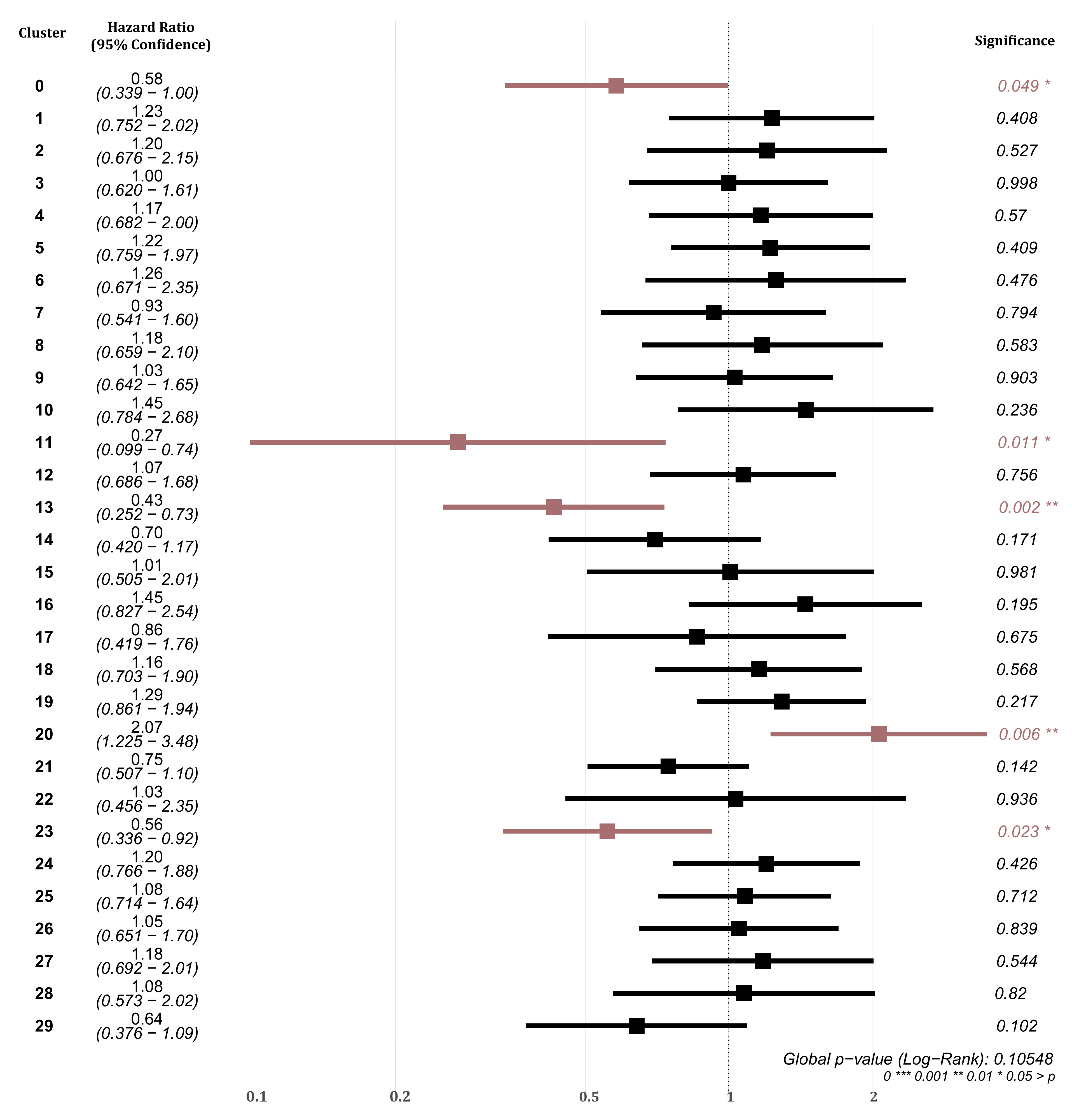}.
  \caption{Multivarite Cox regression comprising all clusters as covariates. Clusters 0, 11, 13, 20, and 23 show significant hazard ratios. Log-Rank Test was used to measure significance of the entire model.}
  \label{fig:multivariate}
\end{figure}

\begin{figure}[H]
  \centering
  \includegraphics[width=\linewidth]{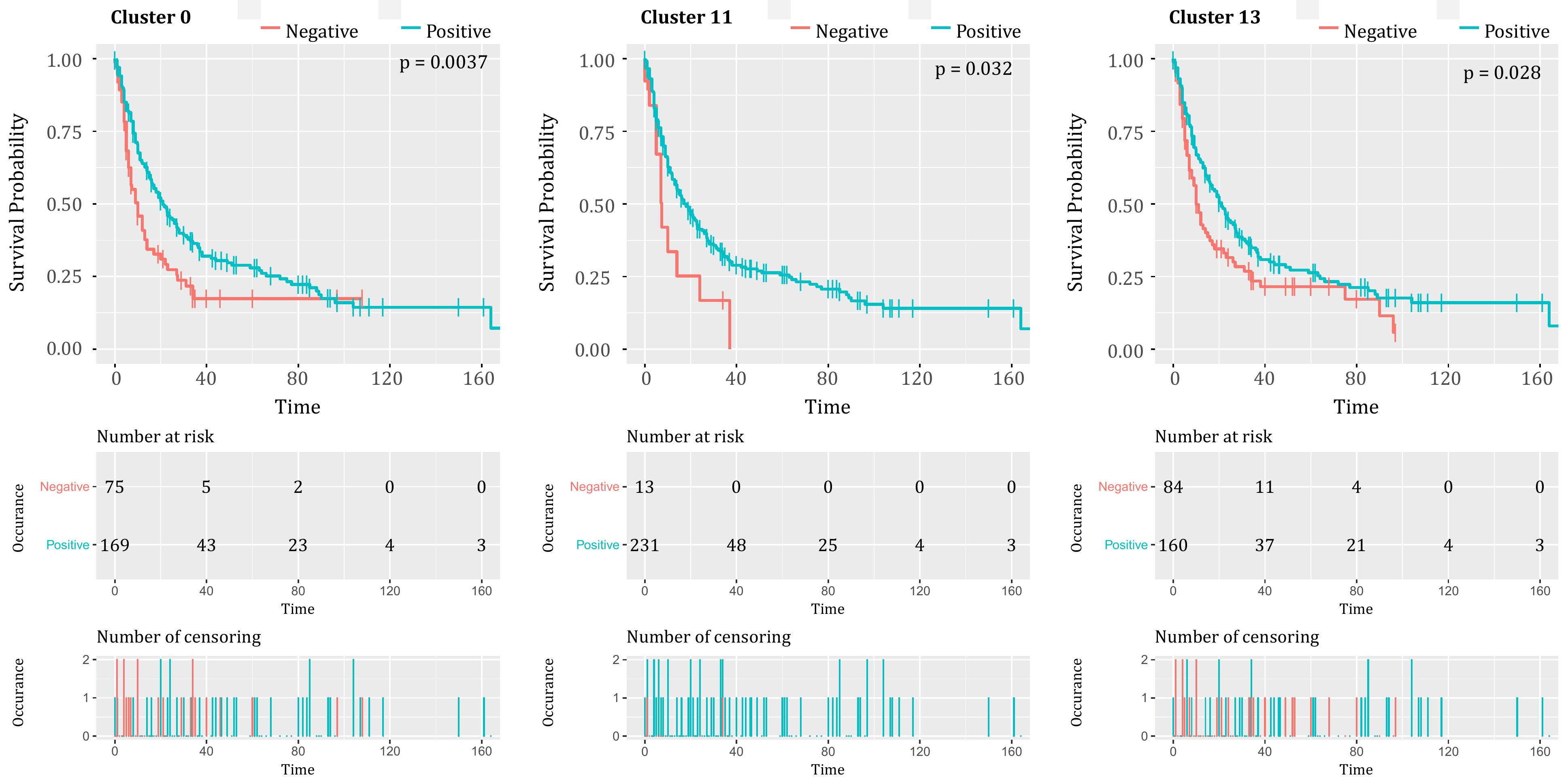}
  \caption{The top panels show Kaplan-Meier survival curves across time (months) for clusters 0, 11, and 13 with reported stratification significance based on the Log-Rank Test. The middle panel shows the amount of samples in each stratified class over time and the bottom panel indicates points at which censored events occur. Each analysis shows a significantly positive prognostic factor for samples positive for the given cluster.}
  \label{fig:unikm}
\end{figure}

For the significant clusters from the univariate analysis, figure \ref{fig:unikm} shows Kaplan-Meier plots for patients stratified into positive and negative groups. A Log-Rank Test p-value of less than 0.05 shows significance in stratification in the estimated survival curves. Each vertical tick indicates a censored event. 

Semi-quantitative histologic analysis of the random 20 tiles selected from the clusters of significance showed that only cluster 0 met criteria for a major feature, consisting of the extracellular matrix component, collagen, specifically arranged in linear fascicles. Collagen was a minor feature for one other cluster (23) and both of these clusters (0, 23) had an decrease in hazard ratio on univariate survival analysis. No tumor histology, as defined in table \ref{tb:elements}, met the criterion as a major feature. One tumor histology was a minor feature of one cluster, clusters 13 had 9 tiles with more than 50 percent solid tumor histology with low nuclear:cytoplasmic ratio and this cluster indicated a decrease in prognostic risk. No other major or minor features were identified.  
 
Although tumor content was not a major or minor feature of most clusters, tumor content of any volume and histologic description was present in 35-75 percent of tiles for each cluster. Major tumor histology types were seen in two clusters: cluster 0  had 4/7 (57 percent) of tiles containing tubular type, and cluster 23 had 7/12 (58 percent) of tiles containing tubular high nuclear:cytoplasmic ratio type.

\begin{table}[H]
\centering
\caption{Hazard ratios ($e^{b_j}$) for prognositically significant clusters $j$ when modelling a univariate Cox regression and their combinations in multivariate models. The values in parenthesis indicate bounds for 95\% confidence intervals based on cumulative hazard.}
  \label{tb:univariate}
\begin{tabular}{@{}llllll@{}}
\toprule
                      & Univariate      & \multicolumn{4}{c}{Multivariate}                                      \\ \midrule \midrule
                      &                 & 0 +11           & 0 +13           & 11 + 13         & 0 + 11 + 13     \\ \midrule
Cluster 0             & 0.618***        & 0.644***        & 0.675**         &                 & 0.725           \\
                      & (0.447 - 0.855) & (0.463 - 0.895) & (0.459 - 0.993) &                 & (0.489 - 1.075) \\ \cmidrule(l){2-6} 
Cluster 11            & 0.515**         & 0.598           &                 & 0.494**         & 0.560*          \\
                      & (0.280 - 0.951) & (0.320 - 1.116) &                 & (0.267 - 0.915) & (0.297 - 1.056) \\ \cmidrule(l){2-6} 
Cluster 13            & 0.750*          &                 & 0.855           & 0.694**         & 0.813           \\
                      & (0.517 - 0.961) &                 & (0.591 - 1.127) & (0.508 - 0.946) & (0.561 - 1.178) \\ \midrule
Wald Test             &                 & 11.36***        & 9.15**          & 9.75***         & 12.51***        \\
Likelihood Ratio Test &                 & 10.19***        & 8.59**          & 8.86**          & 11.37***        \\
Score (Log-Rank) Test &                 & 11.67***        & 9.31***         & 9.99***         & 12.85***        \\ \bottomrule
Note:                 & \multicolumn{5}{r}{Significance codes:  0 ‘***’ 0.001 ‘**’ 0.01 ‘*’ 0.05 ‘.’ 0.1 ‘ ’ 1}
\end{tabular}
\end{table}
 
 \begin{figure}[H]
  \centering
  \includegraphics[width=\linewidth]{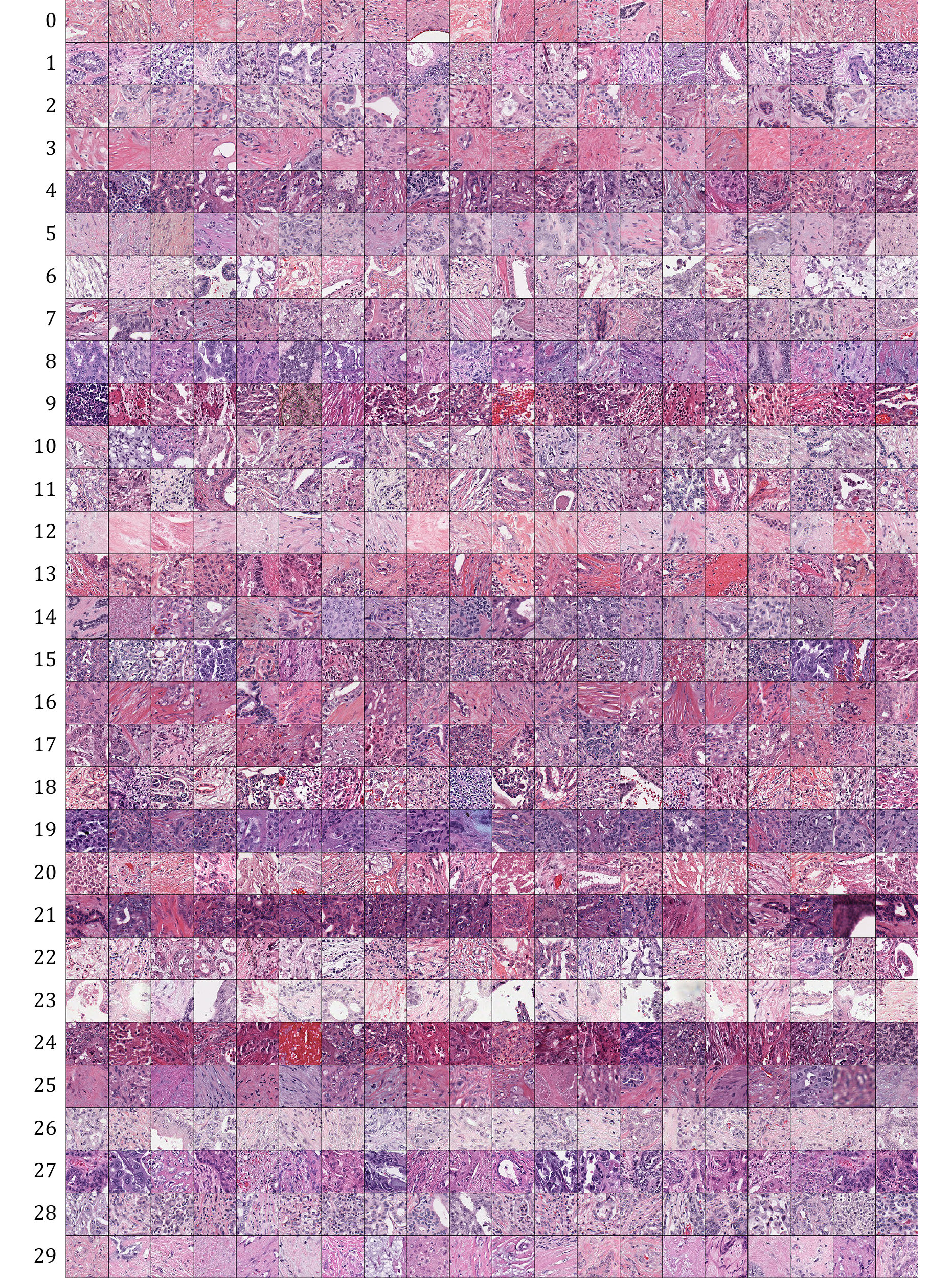}
  \caption{Each row depicts 20 randomly sampled tiles for each cluster.}
  \label{fig:grid}
\end{figure}

\subsection{Conclusion}
\subsubsection{Technical Contribution}
Our model offers a novel approach for identifying histological patterns of potential prognostic significance, circumventing the tasks of tedious tissue labeling and laborious human evaluation of multiple whole slides. As a point of comparison, a recent study showed that an effective prognostic score for colorectal cancer was achieved by first segmenting a slide into eight predefined categorical regions using supervised learning \cite{kather2019predicting}. Methods such as this limit the model to pre-defined histologic components (tumor, non-tumor, fat, debris, etc) and the protocol may not extend to extra-colonic anatomic sites lacking a similar tumor specific stroma interactions \cite{balkwill2012tumor}. In contrast, the design of our model lacks predefined tissue classes, and has the capability to analyze $n$ clusters, thus removing potential bias introduced by training and increasing flexibility in application of the model. 

\subsubsection{Histology}
By semi-quantitative assessment of the histologic components of the tiles in clusters with prognostic significance, we discovered that tumor cells were not a major feature in any cluster, whereas two clusters had connective tissue (stroma) comprised of extracellular matrix (collagen) as a major/minor feature.  


Tumor stroma, the connective tissue intervening between tumor cell clusters in tissue comprised of matrix and collagen, is known to play an integral role in cancer growth, angiogensis, and metastasis, but is not used in tumor grading or prognostication systems, which tend to focus on tumor attributes such as nuclear features, cellular architecture, and invasive behavior  \cite{quail2013microenvironmental, egeblad2010tumors, joyce2009microenvironmental}. Research specifically in ICC has supported the important biological role of tumor associated stroma in tumor progression by analyzing unseen paracrine factors  \cite{terada1994expression, sirica2014desmoplastic, brivio2017tumor}. Recently, a deep learning-based algorithm used tumor associated stroma, not tumor cells, to stratify ductal carcinoma in situ of the breast by grade\cite{bejnordi2018using}. In the present study, we found tumor stroma to be a major/minor feature of two significant clusters which this raises the possibility that the stroma microenvironment could have distinct morphologic characteristics that could be detectable routinely and potentially prognostically significant. 

\subsection{Acknowledgements}
This work was supported by Cycle for Survival, the generous computational support given by the Warren Alpert foundation, and project management from Christina Virgo. Administrative support was provided by Tanisha Daniel, Shanna Guercio, Sarah King, and Patricia Ashby.

Thomas J. Fuchs is a founder, equity owner, and Chief Scientific Officer of Paige.AI.

\bibliographystyle{unsrt}

\end{document}